\documentclass[11pt,a4paper]{article}
\usepackage[hyperref]{arxivjoint}
\usepackage{times}
\usepackage{latexsym}

\usepackage{url}
\usepackage{makecell}
\usepackage{multirow}
\usepackage{multicol}
\usepackage{bm}
\usepackage{algorithm,algpseudocode}
\usepackage{color}
\usepackage{booktabs}
\usepackage{amsmath}
\usepackage{amsfonts}
\usepackage{amssymb}
\usepackage{graphicx}
\usepackage{subfigure}

\aclfinalcopy % Uncomment this line for the final submission

\newcommand*{\affaddr}[1]{#1} % No op here. Customize it for different styles.
\newcommand*{\affmark}[1][*]{\textsuperscript{#1}}
\newcommand*{\email}[1]{\texttt{#1}}

\usepackage{microtype}

\aclfinalcopy % Uncomment this line for the final submission

\title{Jointly Modeling Aspect and Sentiment \\
with Dynamic Heterogeneous Graph Neural Networks}

\author{Shu Liu\affmark[1]\thanks{\ \ Equal Contribution}, Wei Li\affmark[2]\footnotemark[1], Yunfang Wu\affmark[2], Qi Su\affmark[2], Xu Sun\affmark[1,2]\\
\affaddr{\affmark[1]Center for Data Science, Peking University}\\
\affaddr{\affmark[2]MOE Key Lab of Computational Linguistics, School of EECS, Peking University}\\
\email{\{shuliu123, liweitj47, wuyf, sukia, xusun\}@pku.edu.cn}\\
}
\date{}

\begin{document}
\maketitle
\begin{abstract}
Target-Based Sentiment Analysis (\textbf{TBSA}) aims to detect the opinion aspects (aspect extraction) and the sentiment polarities (sentiment detection) towards them. Both the previous pipeline and integrated methods fail to precisely model the innate connection between these two objectives. In this paper, we propose a novel dynamic heterogeneous graph to jointly model the two objectives in an explicit way. Both the ordinary words and sentiment labels are treated as nodes in the heterogeneous graph, so that the aspect words can interact with the sentiment information. The graph is initialized with multiple types of dependencies, and dynamically modified during real-time prediction.
Experiments on the benchmark datasets show that our model outperforms the state-of-the-art models. Further analysis demonstrates that our model obtains significant performance gain on the challenging instances under multiple-opinion aspects and no-opinion aspect situations.
\end{abstract}

\section{Introduction}

% What is this task about?
Target-Based Sentiment Analysis (\textbf{TBSA}) aims to detect the sentiment polarity for specific aspects\footnote{The entities associated with a sentiment is considered as an aspect in \textbf{TBSA} task.} within the text. 
% The significance of the task
For example, in the sentence ``Great service but dreadful food!'', both ``positive'' and ``negative'' sentiment polarities are expressed on different aspects, ``service'' and ``food'' with the opinion words ``great'' and ``dreadful''. 

% Research history and flaws
%   1. separate the tasks
%   2. superficial combination of the two tasks
Traditionally, this problem is solved in two separate steps, namely, \textbf{Aspect Extraction} (\textbf{AE}) and \textbf{Aspect-Based Sentiment Classification} (\textbf{AS}). Although straightforward, this pipeline style solution ignores the innate connection between the two objectives, resulting in error propagation. Recently, \citet{wang2018towards}, \citet{li2019unified} propose to use the connection between the two tasks by unified tags or combination of feature vectors, which remain on modeling the joint features in superficial ways. The connection between two tasks can only be modeled in the shared layers, which is an implicit and black-box method. Furthermore, information between task-specific layers can only be transmitted via the shared layers, which is not explicit nor direct.

% Observations and motivations
% 实体不一定携带情感
When studying the \textbf{TBSA} task, we observe two facts. \textbf{(1) Not all the entities in the text are associated with a sentiment}. For instance, ``\textit{I haven't used it for anything high tech yet, but I love it already}'', the sentiment target is pronoun ``it'' rather than the entity ``tech''. However, the pipeline method suffers from improperly labeling all the entities (such as ``tech'') without being aware of the sentiment information.
% 多实体情况需要对句子进行多次迭代
\textbf{(2) Many cases in \textbf{TBSA} involve multiple aspects, which are often related to each other on both the aspect sense and the sentiment sense}. For instance, ``\textit{I like coming back to Mac OS but this laptop is lacking in speaker quality compared to my \$400 old HP laptop.}'', the opinion words ``like'' and ``lacking'' are indicators conveying the corresponding sentiment polarities of aspects ``Mac OS'' and ``speaker quality''. In addition, the word ``but'' indicates the opposite sentiments of the two aspects. The previous works can not precisely model such complex relationships in an explicit and effective way.

To address the above challenges, we propose a dynamic heterogeneous graph mechanism to represent the complex dependencies among words and sentiment labels.
To model the connections between different sentiments and aspects, we consider both the sentiment labels and the ordinary words as nodes in the graph. The two kinds of nodes are connected by real-time prediction of sentiments toward words during the iterative predicting process. Specifically, the edges with high prediction confidence are firstly built, which makes the sub-task predicted information explicitly encoded in the graph. In the next prediction iteration, the model predicts other aspects and sentiment based on the modified graph. In such settings, the graph becomes a dynamic heterogeneous graph, which provides an explicit way to model the connection between the two objectives other than the implicit modeling method of previous works.

\begin{figure}
\centering
\includegraphics[trim=0cm 0cm -0.5cm 0cm, width=0.5\textwidth]{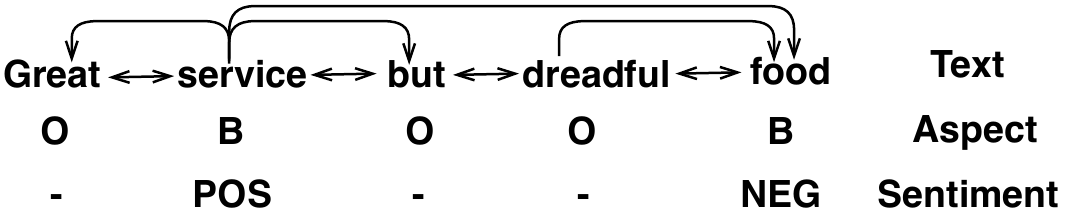}
\vspace{-0.2in}
\caption{An illustration of the \textbf{TBSA} task. The top arrows are the illustration of syntactic and positional dependency between words. The bottom shows the target labels of the sentence, including the aspect labels (BIO) and sentiment labels (POS, NEG, NEU).}
\label{fig:graph}
\vspace{-0.15in}
\end{figure}

In addition, to capture the long-range dependency between aspect words and sentiment opinion words~\citep{huang2019syntax, sun-etal-2019-aspect}, we propose to connect the word nodes with syntactic and positional dependencies, illustrated in Figure~\ref{fig:graph}. The combination of heterogeneous node types and the usage of long-range dependency not only shortens the distance between aspect words and sentiment opinion words, but also help build the path between opinion word and sentiment labels. 
To represent multiple types of relations in the heterogeneous graph, we propose a Heterogeneous Gated Graph Neural Network to encode the graph.

% Experiment result
We do experiments on three benchmark SemEval-Task datasets and conduct detailed analysis. Experiment results show that our model beats state-of-the-art models with the same amount of training data by a big margin. Even compared with models using much external data, our model achieves very competitive results.

% Contributions
We conclude our contributions as follows:
\begin{itemize}
\item We propose to explicitly model the connection between the two sub-tasks \textbf{AE} and \textbf{AS} in \textbf{TBSA} with dynamic heterogeneous graph mechanism that include both words and sentiment labels as nodes. The iterative modification of graph bridges the words and sentiments, which makes the latter prediction aware of the previous predicted sentiments. 
\item We propose to model the long-range dependencies with syntactic and positional edges and propose Heterogeneous Gated Graph Neural Network to encode the heterogeneous dependencies. The combination of heterogeneous node types and the usage of long-range dependency not only shorten the distance between aspect words and opinion words, but also help build the path between opinion words and sentiment labels.
\item Experiments show that our model outperforms the state-of-the-art models. Extensive analysis proves the advantages of our model: \textbf{\uppercase\expandafter{\romannumeral1}}. providing the explicitly interpretable connection between sub-tasks via the graph modification, \textbf{\uppercase\expandafter{\romannumeral2}}. excluding aspects without opinion, \textbf{\uppercase\expandafter{\romannumeral3}}. boosting the performance on the multi-opinion-aspects instances.
\end{itemize}

\section{Proposed Method}

\begin{figure*}
\centering
\includegraphics[trim=0cm 0cm 0cm 0.2cm, width=1\textwidth]{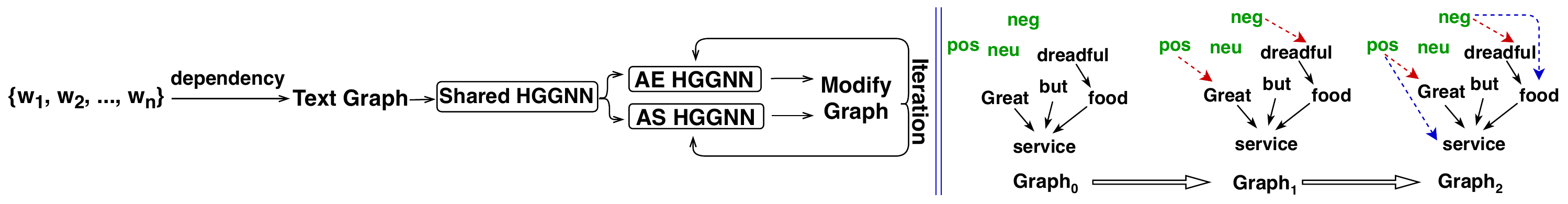}
\vspace{-0.3in}
\caption{An illustration of our model structure. The {\color{red}red dot} arrow and the {\color{blue}blue dot} arrow in the graph denote the first iteration and the second iteration ``sentiment'' edge we built based on the real-time prediction. $\textbf{Graph}_{\textbf{l}}$ denotes the graph during the $l$-th iteration. \{{\color{green}pos, neg, neu}\} are sentiment nodes.}
\label{fig:model}
\end{figure*}

%In this section, we first briefly introduce the task formulation of TBSA. Then,  we introduce our proposed Heterogeneous Gated Graph Neural Network (HGGNN), which is the basic structure of our model.

%Next, we describe our model illustrated in Figure~\ref{fig:model} in details. Firstly, we initialize a text graph utilizing syntactic and positional dependency. Secondly, we use HGGNN to learn the fundamental context representation shared by the two objectives. Thirdly, in AS and AE sub-tasks, we propose a novel dynamic heterogeneous graph setting to predict the different sentiments and different aspects (right part of Figure~\ref{fig:model}), where the sentiment labels are treated as nodes as well as the words. 

In this section, we first briefly introduce the formulation of \textbf{TBSA} task. Then, we describe how to construct the dynamic heterogeneous graph in two steps: (1) text graph initialization (2) dynamic heterogeneous graph modification. Next, we show how to encode the heterogeneous graph with our proposed Heterogeneous Gated Graph Neural Network (HGGNN) model. Lastly, we show how to train the model for the two objectives, \textbf{AE} and \textbf{AS}. A brief illustration is shown in Figure~\ref{fig:model}.

\subsection{Task Formulation}
We treat the \textbf{TBSA} problem as a combination of \textbf{AE} and \textbf{AS} objectives. \textbf{AE} is formulated as a sequence tagging problem with the ``BIO'' labels. \textbf{AS} is also formulated as a sequence tagging problem with labels \{POS, NEG, NEU\} labels. Figure~\ref{fig:graph} is an example illustrating the ground truth labels.

%\subsection{Modeling Word Dependency with Shared Graph Representation}\label{sec:shared_ggnn}

\subsection{Dynamic Heterogeneous Graph Mechanism}

\noindent \textbf{Text Graph Initialization:}\\
We initialize the graph as a directed graph using  syntactic and positional dependencies among word nodes. 
The text graph has three different types of edges: unidirectional syntactic dependency edges ``to'' (denoted as $\mathcal{N}^{t}$) and ``from'' (denoted as $\mathcal{N}^{f}$), bidirectional ``position'' edges based on the local window  (denoted as $\mathcal{N}^{p}$). Dependency parsing is used to detect the syntactic relations, which helps detect sentiment features related to the aspect words.
\\[0.05in]
\noindent \textbf{Dynamic Heterogeneous Graph Modification:}\\
Based on the observation that two objectives and the sentiments of different aspects are not independent, we propose to model both the text and sentiment labels as nodes in a heterogeneous graph. In order to use the information of different types, we propose to use a fourth type of edge ``sentiment'' in addition to the other three relations described above. This kind of edge is built based on real-time predictions in \textbf{AS}.

In such settings, the graph becomes a dynamic heterogeneous graph. The graph is heterogeneous as there are two types of nodes and four types of edges. The graph is dynamic because the ``sentiment'' edges between word and sentiment nodes are dynamically built and modified during the real-time prediction process rather than fixed. The iterative modification process on graph is shown in the right part of Figure~\ref{fig:model}.
which is formulated as follows:

\begin{small}
\begin{equation}
\begin{aligned}
p(y|\Theta, X) &= p(y_l|\Tilde{y}_{l-1}, \cdots, \Tilde{y}_{1}, \Theta, X) \\
& \times p(y_{l-1}|\Tilde{y}_{l-2}, \cdots, \Tilde{y}_{1}, \Theta, X) \\
& \times \cdots \times p(y_1|\Theta, X)
\end{aligned}
\end{equation}
\end{small}
where $y$ means the sentiment polarity of each token, $y_l$ means the  sentiments in the $l$-th iteration, $\widetilde{y_i}$ represents the sentiments with high confidence in the $i$-th iteration . $\Theta$ and $X$ mean the parameters and the inputs respectively.

\begin{algorithm}[t]
\small
\caption{DHG Mechanism}
\label{alg:dynamic_heterogeneous_graph}
\begin{algorithmic}
\Require{\\
$\bm{h}_{\textit{senti}_s}:$ node embedding of sentiment $s$\\
Graph $\mathcal{G}^0:$ initial graph \\
$\bm{x}_i:$ embedding of word $i$\\
$\bm{m}_{i}^0:$ initial hidden vector of aspect-part\\
$\bm{n}_{i}^0:$ initial hidden vector of sentiment-part\\
$\epsilon$, $\textit{times}$} 
\Ensure{\\
$\bm{m}_{i}^{\textit{times}}:$ final hidden vector of aspect-part\\
$\bm{n}_{i}^{\textit{times}}:$ final hidden vector of sentiment-part
} 
\Statex
\For{$l \in 1$ to $\textit{times}$}                    
    \State {$\bm{m}_{i}^l =  \textit{HGGNN}_{\textbf{AE}}(\bm{m}_{i}^{l-1}, \bm{x}_i, \mathcal{G}^{l-1})$}
    \State {$\bm{n}_{i}^l = \textit{HGGNN}_{\textbf{AS}}(\bm{n}_{i}^{l-1}, \bm{x}_i, \mathcal{G}^{l-1})$}
    \State {$prob_{is}^l = \textit{softmax}(\textit{sim}(\bm{n}_i^l, \bm{h}_{senti_s}))$}
    \For{word $i$, sentiment $s$} 
        \If {$prob_{is}^l > \epsilon$}
            \State Graph $\mathcal{G}^l$ $\gets$ Graph $\mathcal{G}^{l-1}$
            \State link the word node $i$ with the sentiment node $s$. 
        \EndIf
    \EndFor
    \State Drop some ``sentiment'' edges.
\EndFor
\end{algorithmic}
\end{algorithm}
Each time we predict the sentiment of an aspect, we add a ``sentiment'' edge between the aspect word node and the sentiment label node. The sentiments with high confidence are first bridged, which make the predicted information explicitly revealed in the graph. When the graph evolves because of the prediction, we update the graph representation based on the new graph structure. By doing so, the current prediction procedure is aware of the historical prediction results, that is, the sentiments of other aspects. This way of organizing the graph makes our method able to model the connection among different sentiments and different aspects. The dynamic process is shown in algorithm~\ref{alg:dynamic_heterogeneous_graph}. Note that after each iteration, we drop some ``sentiment'' edges to make the degree of sentiment nodes meet their distributions in the training set. This operation prevents the predominant predictions on specific sentiment labels.

The predictions are not accurate enough at the early training stage. To prevent slow convergence, we adopt the teacher forcing training method. Specifically, we add some of the edges between word and sentiment according to the ground truth labels at the beginning of training. At each training epoch, we modify the graph based on the predicted labels with high confidence probability $\widetilde{p}$, otherwise we adopt the teacher forcing training method. We define $\widetilde{p}$ following \citet{zhang2019bridging}:
\begin{equation}
\begin{small}
    \widetilde{p} = 1-\frac{\mu}{\mu+ \exp(epoch/\mu)}
\end{small}
\end{equation}
In teacher-forcing, we randomly dropout 80\% ground truth to prevent our model from copying the ground truth straightly.

\subsection{Heterogeneous Gated Graph Neural Network}

Gate mechanism largely enhances the expression ability of GNN models and is able to release the over-smoothing problem\footnote{Over-smoothing problem means all nodes will converge to very similar values when stacking multiple layers, which will cause information loss~\citep{li2018deeper,chen2019measuring}.}~\citep{li2015gated, peng2017cross, zhang2018sentence}.
We propose a Heterogeneous Gated Graph Neural Network (HGGNN) to encode the heterogeneous dependency relationships among words that adapts the gate mechanism into heterogeneous graph area. %Gated Graph Neural Network is proposed in~\citet{li2015gated}, which has a strongly expression and the ability to release the over-smoothing problem\footnote{Over-smoothing problem means all nodes will converge to very similar values when stacking multiple layers, which will cause information loss~\citep{li2018deeper,chen2019measuring}.}. We extend the model on the heterogeneous graph as the words has multiple relationships between each other.

For each relationship (i.e. edge), we first get the neighbour vector of each word:

\vspace{-6pt}
\begin{small}
\begin{align}
\bm{a}_i^t = A_i^{T}[\bm{h}_1^{t-1}, \cdots,\bm{h}_{|\mathcal{V}|}^{t-1}]^T+\bm{k}\notag\\
\bm{b}_i^t = B_i^{T}[\bm{h}_1^{t-1}, \cdots,\bm{h}_{|\mathcal{V}|}^{t-1}]^T+\bm{k}\label{eq:neighbor_eq}\\
\bm{c}_i^t = C_i^{T}[\bm{h}_1^{t-1}, \cdots,\bm{h}_{|\mathcal{V}|}^{t-1}]^T+\bm{k}\notag
\end{align}
\end{small}
where the matrix $A_i$, $B_i$, $C_i$ determine how word node $i$ communicate with each other in the graph for each relationship. $t$ is the layer index. Then, we take the embeddings of the nodes and the graph structure as input and outputs the hidden representation of each node. Gated Recurrent Units~\citep{cho2014learning} is used to dynamically decide which part of information should be transmitted to upper layers to alleviate the over-smoothing problem:

\vspace{-6pt}
\begin{small}
\begin{align}
\bm{z}_i^t &= \sigma(W_a^z\bm{a}_i^t+W_b^z\bm{b}_i^t+W_c^z\bm{c}_i^t+V^z\bm{x}_{i} +U^z \bm{h}_i^{t-1}) \notag\\ 
\bm{r}_i^t &= \sigma(W_a^r\bm{a}_i^t+W_b^r\bm{b}_i^t+W_c^r\bm{c}_i^t+V^r\bm{x}_{i} +U^r \bm{h}_i^{t-1}) \notag\\
\widetilde{\bm{h}}_i^{t} &= \tanh(W_a\bm{a}_i^t+W_b\bm{b}_i^t+W_c\bm{c}_i^t+V\bm{x}_{i}+ \label{eq:tailor_ggnn}\\
&\ \ \ \ \ \ \ \ \ \ \ \ \ \ \ U(\bm{r}_i^t \odot \bm{h}_i^{t-1})) \notag\\
\bm{h}_i^{t} &= (1-\bm{z}_i^t) \odot \bm{h}_i^{t-1} + \bm{z}_i^t \odot \widetilde{\bm{h}}_i^{t} \notag
\end{align}
\end{small}
where $\bm{x}_i$ and $\bm{h}_i$ are the embedding and the hidden state of node $i$. $t$ is the layer index. $\bm{z}$ and $\bm{r}$ mean update gate and reset gate, respectively.

\begin{table*}
\centering
\small
\begin{tabular}{c|ccccc|ccccc}\hline
& \multicolumn{5}{c|}{Train} & \multicolumn{5}{c}{Test} \\
Datasets & Aspect & Pos & Neg & Neu & Conf & Aspect & Pos & Neg & Neu & Conf \\
\hline
Res14 & 3,692 & 2,160 & 804 & 637 & 91 & 1,132 & 728 & 195 & 195 & 14 \\
Lt14 & 2,373 & 994 & 870 & 464 & 45 & 654 & 341 & 128 & 169 & 16 \\
Res15 & 1,199 & 902 & 252 & 34 & 11 & 542 & 319 & 179 & 27 & 17 \\
\hline
\end{tabular}
\caption{Dataset statistics. ``Aspect'' means the number of entities with sentiment. ``Pos, Neg, Neu, Conf'' represent the sentiment polarities \textit{positive, negative, neutral and conflict} respectively.}\label{tab:statistics}
\vspace{-0.12in}
\end{table*}

\subsection{Training and Inference}

\noindent\textbf{AE Objective:} At the top of the ``\textbf{AE} HGGNN'', we add a CRF layer to calculate the conditional probabilities of aspect label sequences as follows:

\vspace{-6pt}
\begin{small}
\begin{align}
P(\mathbf{\widehat{Y}}^{\textbf{AE}}|\bm{m},\mathbf{W_s},\mathbf{b_s}) = \frac{\prod_{i=1}^n\phi_i(\bm{y}_{i-1},\bm{y}_i\,\bm{m})}{\sum_{\mathbf{Y'}}\prod_{i=1}^n\phi_i(\bm{y}'_{i-1},\bm{y}'_i\,\bm{m})}\\
\phi_i(\bm{y}_{i-1},\bm{y}_i,\bm{m})=\exp(\mathbf{W}^{\bm{y}_{i-1},\bm{y}_i}_s m_i+b_s^{\bm{y}_{i-1},\bm{y}_i})
\end{align}
\end{small}
where $\bm{m}$ represents the final hidden vector of graph in aspect-level part, $\mathbf{\widehat{Y}}^{\textbf{AE}}$ means the aspect labels (B, I, O) of each tokens. $\mathbf{W_s}$ and $\mathbf{b_s}$ are transition parameters to be trained. Then the loss function of \textbf{AE} objective is

\begin{small}
\begin{equation}
L_{\textbf{AE}} = \frac{1}{N_a}\sum_{i=1}^{N_a}\frac{1}{n_i}\sum_{j=1}^{n_i}l(\mathbf{Y}_{i,j}^{\textbf{AE}}, \mathbf{\widehat{Y}}^{\textbf{AE}}_{i,j})
\end{equation}
\end{small}
where $N_a$ denotes the total number of sentences in the training set, $n_i$ denotes the number of tokens of the $i$-th sentence, $\mathbf{Y}_{i, j}^{\textbf{AE}}$ and $\mathbf{\widehat{Y}}^{\textbf{AE}}_{i, j}$ denote the ground truth aspect label and the predicted label of the $j$-th token in the $i$-th sentence respectively. $l(,)$ is the cross-entropy loss function.
\\[0.05in]
\noindent\textbf{AS Objective:} At the top of the ``\textbf{AS} HGGNN'', we add an MLP layer to compute the probabilities of sentiment label:
\begin{equation}
\begin{small}
P(\mathbf{\widehat{Y}}^{\textbf{AS}}|\bm{n}) = \textit{softmax}(f(\bm{W}\bm{n}+\bm{b}))
\end{small}
\end{equation}
To activate the DHG mechanism, we add the prediction based on the similarities between word nodes and sentiment nodes:

\begin{small}
\begin{equation}
P(\mathbf{\widetilde{Y}}^{\textbf{AS}}|\bm{n}, \bm{h}_{senti}) = \textit{softmax}(\left<\bm{n}, \bm{h}_{senti}\right>)
\end{equation}
\end{small}
where $\bm{n}$ is the final hidden vector of the sentence in sentiment-level part. $\left<, \right>$ is the inner product operation. $\mathbf{\widehat{Y}}^{\textbf{AS}}_{i, j}$ and $\mathbf{\widetilde{Y}}^{\textbf{AS}}_{i, j}$ are the predicted sentiment label. Then the loss function of \textbf{AS} task is:

\vspace{-6pt}
\begin{small}
\begin{equation}
\begin{aligned}
L_{\textbf{AS}} =& \frac{1}{2} \frac{1}{N_a}\sum_{i=1}^{N_a}\frac{1}{n_i}\sum_{j=1}^{n_i}l(\mathbf{Y}_{i,j}^{\textbf{AS}}, \mathbf{\widehat{Y}}^{\textbf{AS}}_{i,j}) +\\
& \frac{1}{2} \frac{1}{N_a}\sum_{i=1}^{N_a}\frac{1}{n_i}\sum_{j=1}^{n_i}l(\mathbf{Y}_{i,j}^{\textbf{AS}}, \mathbf{\widetilde{Y}}^{\textbf{AS}}_{i,j})
\end{aligned}
\end{equation}
\end{small}
where $\mathbf{Y}_{i, j}^{\textbf{AS}}$ is the ground truth sentiment label.
\\[0.05in]
\noindent\textbf{Integrated Objective: } The total loss is calculated as the sum of the two sub-objectives:
\begin{equation}
\begin{small}
    L = L_{\textbf{AE}} + \lambda L_{\textbf{AS}}
\end{small}
\end{equation}
where $\lambda$ is a hyper-parameter to balance the loss of two objectives (we set $\lambda=1$).

During inference, we get the extracted aspect words and the sentiment results on them. To tackle sentiment inconsistency, where one aspect is attached with multiple different sentiment tokens, we average the sentiment probability of every token within an aspect to predict the polarities.

\section{Experiment}
\subsection{Datasets}

We conduct experiments on three widely used benchmark datasets of SemEval 2014~\citep{pontiki-etal-2014-semeval} and 2015~\citep{pontiki2015semeval}. The statistics of these datasets is shown in Table~\ref{tab:statistics} (denoted as Res14, Lt14 and Res15). All these datasets contain the ground truth labels of both the target aspect and their sentiment polarities. We follow the standard train-test set split, and randomly sample 20\% of the training data as development set. We ignore aspect terms with ``conflict'' sentiment polarities following the previous works.

\subsection{Baselines}
We compare our model with state-of-the-art \textbf{pipeline} and \textbf{integrated} baselines.
\\[0.13in]
\noindent\textbf{Pipeline:} Pipeline models consist of two parts: \\(\textbf{\uppercase\expandafter{\romannumeral1}}). \textbf{AE} models include \textbf{CMLA}~\citep{wang2017coupled} and \textbf{DECNN}~\citep{xu2018double}. \textbf{CMLA} use a novel end-to-end network with coupled multi-layer attentions for aspect-opinion co-extraction. \textbf{DECNN} is a multi-layer CNN  with double embeddings.\\ (\textbf{\uppercase\expandafter{\romannumeral2}}). \textbf{AS} models include \textbf{ATAE}~\citep{wang2016attention} and \textbf{TransCap}~\citep{chen2019transfer}. \textbf{ATAE} applies an attention-based LSTM structure, which incorporates the aspect embedding as input. TransCap applies an aspect routing approach with 60,000 external sentence-level sentiment data from Yelp and Amazon. We select the above models to construct 4 pipeline baselines: \textbf{CMLA-ATAE}, \textbf{CMLA-TransC}, \textbf{DECNN-ATAE}, \textbf{DECNN-TransCap} (denoted as C-A, C-T, D-A and D-T).
\\[0.13in]
\noindent\textbf{Integrated:} \textbf{E2ETBSA}~\citep{li2019unified} propose a unified tagging model to represent the constrained transitions from target boundaries to target sentiments. Opinion lexicon is further used to enhance the aspect extraction component. \textbf{IMN-d}~\citep{he2019interactive} propose an interactive multi-task learning network with a message passing architecture, which also extends \textbf{IMN-d} model to \textbf{IMN} model with 60,000 extra document-level data from Yelp and Amazon.

\subsection{Settings}
We use the tokenized datasets released in~\citet{he2019interactive} for a fair comparison. Text node embeddings are initialized with the concatenation of the general-purpose embeddings\footnote{Pre-trained 300 dimension Glove vectors~\citep{pennington-etal-2014-glove}.} and domain embeddings\footnote{Pre-trained 100 dimension vectors on a domain-specific corpus using fastText released in~\citet{xu2018double}.}~\citep{xu2018double,he2019interactive}. 
%The embeddings of the out-of-vocabulary words are sampled from the uniform distribution $\mathcal{U}(-0.25, 0.25)$~\citep{kim2014convolutional}. 
For the graph structure, we use Stanford Dependency Parser~\citep{chen-manning-2014-fast} to obtain syntactic dependency. We set the window size to 3 to obtain positional neighbours. 
%We initialize the weight matrices by the uniform distribution $\mathcal{U}(-0.1, 0.1)$. All bias matrices are initialized to 0. 
We set the layers of $\textit{HGGNN}_{\textbf{Shared}}$, $\textit{HGGNN}_{\textbf{AE}}$, $\textit{HGGNN}_{\textbf{AS}}$ to 3. On \textbf{Res14} datasets the \textbf{DHG} iteration times is set to 3, while 2 on \textbf{Lt14} and \textbf{Res15} datasets. We set the \textbf{DHG} threshold value $\epsilon$ to $0.75$. We use Adam optimizer~\citep{KingmaBa2014} with $0.0001$ learning rate, and train our model with batch size of 32. We apply $0.5$ dropout regularization~\citep{dropout} and clip the gradients to the maximum norm of 1.0. All hyper-parameters are tuned on the development set.

\begin{table*}
\centering
\begin{tabular}{cc|cccc|cc|c}
\hline
\multicolumn{2}{c|}{} & \multicolumn{4}{c|}{Pipeline Models} & \multicolumn{2}{c|}{Integrated Models} & \multirow{2}{*}{\textbf{Our model}}\\
 & & C-A* & D-A* & C-T & D-T & E2ETBSA* & IMN-d* & \\
\hline
\multirow{4}{*}{Res14} 
& F-a & 82.45\% & 83.33\% & 82.71\% & 83.52\% & 83.12\% & 83.89\% & \textbf{84.82\%} \\
& acc-s & 77.46\% & 77.63\% & 79.65\% & 79.65\% & 79.06\% & 80.69\% & \textbf{80.91\%} \\
& F-s & 68.70\% & 70.09\% & 70.91\% & 70.91\% & 68.77\% & \textbf{72.09\%} & 71.37\% \\
& F-all & 63.87\% & 64.32\% & 65.89\% & 66.54\% & 65.94\% & 67.27\% & \textbf{68.91\%} \\
\hline
\multirow{4}{*}{Lt14} 
& F-a & 76.80\% & 80.28\% & 77.50\% & \textbf{80.56\%} & 77.67\% & 78.43\% & 80.12\% \\
& acc-s & 70.25\% & 69.98\% & 72.83\% & 72.83\% & 71.72\% & 72.49\% & \textbf{74.51\%} \\
& F-s & 66.67\% & 66.20\% & 68.83\% & 68.83\% & 68.36\% & 69.71\% & \textbf{70.48\%} \\
& F-all & 53.68\% & 55.92\% & 56.46\% & 58.43\% & 55.95\% & 57.13\% & \textbf{59.61\%} \\
\hline
\multirow{4}{*}{Res15} 
& F-a & 68.55\% & 68.32\% & 68.61\% & 68.40\% & 68.79\% & 70.35\% & \textbf{70.93\%} \\
& acc-s & 81.03\% & 80.32\% & 81.84\% & 81.84\% & 80.96\% & 81.86\% & \textbf{82.53\%} \\
& F-s & 58.91\% & 57.25\% & 66.10\% & 66.10\% & 57.10\% & 56.88\% & \textbf{68.30\%} \\
& F-all & 54.79\% & 55.10\% & 56.21\% & 55.95\% & 55.45\% & 57.86\% & \textbf{58.37\%} \\
\hline
\end{tabular}
\caption{Comparison between our proposed model and the baselines without external data. * means that the result is copied from the previous paper. Otherwise, average results over 5 runs with different random seeds are reported. The evaluation methods are described in section \ref{sec:evaluation}.}\label{tab:sota}
\vspace{-0.1in}
\end{table*}

\subsection{Evaluation Metrics}\label{sec:evaluation}

We employ four metrics for evaluation. we use \textbf{F-all} to measure the performance of \textbf{TBSA}. F-all is similar to F-score, where the \{aspect, sentiment\} result is considered as correct only when both elements are correct. Because \textbf{F-all} is the metric we care most, we train our model for 200 epochs, and save the checkpoint with the best F-all on development set for evaluation. For \textbf{AE} task, we use F-score to measure the performance of aspect extraction (denoted as \textbf{F-a}). For \textbf{AS} task, we use accuracy and macro F-score to evaluate (denoted as \textbf{acc-s} and \textbf{F-s}). 

\subsection{Results}

Table~\ref{tab:sota} reports the results of our model and baselines. For a fair comparison, the results reported here do not involve any external data other than the SemEval datasets. We can observe that the carefully-designed integrated models generally beat the pipeline models. However, since these previous integrated models only make use of the superficial connections between the two sub tasks, our model outperforms the state-of-the-art baselines by 1.64\%, 2.48\% and 0.51\% on F-all metric in all the three datasets. This shows the effectiveness of our method in modeling the interaction between two sub-tasks with the shared graph representation and the explicit information exchange during the iterative prediction and graph modification. 

Furthermore, our model achieves best results on most of the tasks regarding to the sub-task metrics, F-a, acc-s and F-s, 
which validates that the results of the individual sub-task can be mutually promoted in our model. Not only the predicted aspect labels can benefit the prediction of sentiment, but the iteratively built edges can in turn help the aspect prediction process. This is because that if the predicted aspects are not connected with any sentiment label nodes, they will not be given aspect labels in the following prediction iterations.

One thing that should be noted is that even compared with models using much external data (\textbf{TransCap} and \textbf{IMN} use 30,000 external sentence-level sentiment data in each domain), our model outperforms these models without any external data. The state-of-the-art F-all scores reported by \textbf{IMN} on Res14, Lt14 and Res15 are 68.71\%, 58.04\% and 58.18\%, which are lower than the results of our model 68.91\%, 59.61\% and 58.37\%.

\section{Analysis}

%In this section, we extend our experiments to analyse our model in detail.

\subsection{Ablation Study}

Table~\ref{tab:ablation} reports the results of ablation study on the development and test set. To investigate the impact of each component, we remove one component at a time from our model and study the effects. 

First, we remove the \textbf{DHG mechanism} (see Algorithm~\ref{alg:dynamic_heterogeneous_graph}) from the model. That is to say, we directly predict the aspect and sentiment labels after getting the graph representation from the shared HGGNN. The result shows without DHG, the joint-result performance drops in all three datasets, which suggests that the interaction between the two sub-tasks indeed provide useful information to the prediction. The DHG mechanism, which iteratively bridges the word nodes and sentiment nodes, enables the latter prediction process to be aware of the previously predicted sentiment of other aspects. 

\begin{table}[t]
\centering
\begin{tabular}{l|ccc}
\hline
\textbf{Dev Set} & Res14 & Lt14 & Res15 \\
\hline
 \textbf{Our Model} & \textbf{62.81\%} & \textbf{57.40\%} & \textbf{66.01\%} \\
 \hline
 \textbf{w.o.} DHG & 61.34\% & 56.08\% & 65.32\% \\
 Syntax $\rightarrow$ PMI & 60.93\% & 55.31\% & 64.98\% \\
\hline
\hline
\textbf{Test Set} & Res14 & Lt14 & Res15 \\
\hline
 \textbf{Our Model} & \textbf{68.91\%} & \textbf{59.61\%} & \textbf{58.37\%} \\
  \hline
 \textbf{w.o.} DHG & 67.77\% & 58.28\% & 57.96\% \\
 Syntax $\rightarrow$ PMI & 66.93\% & 57.56\% & 56.88\% \\
\hline
\end{tabular}
\caption{Ablation study on development and test set. ``\textbf{w.o.} DHG'' means without DHG mechanism, ``Syntax $\rightarrow$ PMI'' means using
PMI edges to replace syntactic edges.}\label{tab:ablation}
\vspace{-0.2in}
\end{table}

We further remove the \textbf{syntax} dependency by replacing the syntax edges with co-occurrence edges built by \textbf{PMI} values\footnote{We compute the PMI value between word $i$ and $j$, and build the co-occurrence edge if value $>$ 0~\citep{yao2019graph}.} to prevent sparsity. We can observe that the performances decline consistently. This testifies that syntax information is helpful to the performance of joint \textbf{TBSA}, because it directly bridges aspect words with other key words.

\subsection{Performance on Instances with Multiple Opinion Aspects}

We argue that DHG mechanism in our model can make the latter prediction process aware of the previously predicted sentiment of other aspects. Therefore, the model can achieve better performance on the \textbf{multi-opinion-aspects} sentences. To prove the effectiveness on this sort of instances, we select the sentences with multiple opinion aspects from the test set. The statistics are shown in Table~\ref{tab:test_statistics}. 

\begin{table}[H]
\centering
\small
\begin{tabular}{c|c|c|c}\hline
& Res14 & Lt14 & Res15 \\
\hline
\#instance & 800 & 800 & 685 \\
\#multi-op-as & 316 & 156 & 107 \\
\#no-op-as & 194 & 378 & 284 \\
\hline
\end{tabular}
\caption{Number of instances, number of multi-aspects and number of no-aspects instances in test set.}\label{tab:test_statistics}
\vspace{-0.1in}
\end{table}

\begin{figure*}
\centering
\subfigure[]{
\begin{minipage}[t]{0.3\linewidth}
\includegraphics[trim=2cm -1cm -1.5cm 2cm, width=2in]{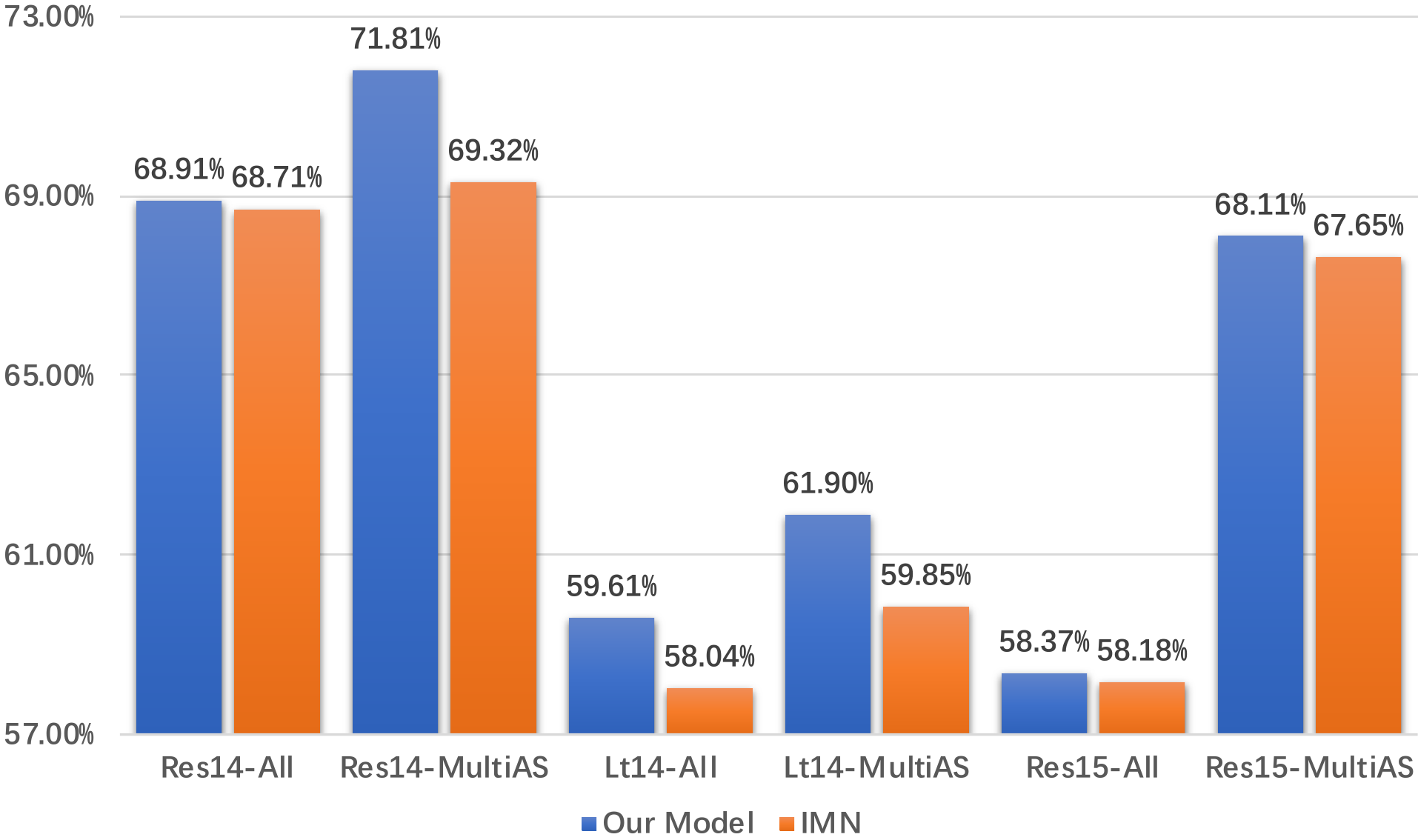}
\vspace{-0.3in}
\label{fig:compare1}
\end{minipage}%
}%
\subfigure[]{
\begin{minipage}[t]{0.3\linewidth}
\centering
\includegraphics[trim=-1.5cm -1cm 0.5cm 1cm, width=2in]{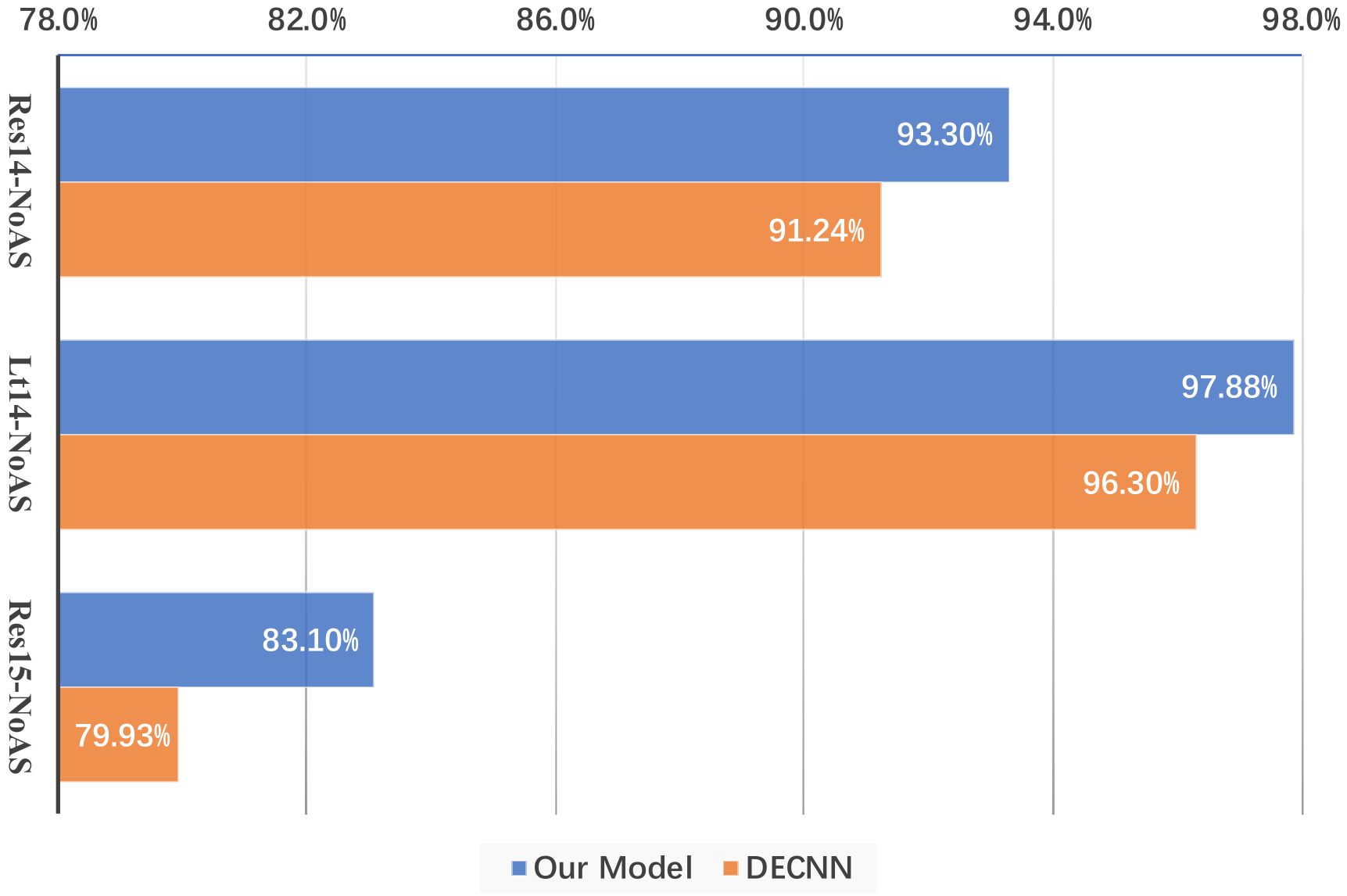}
\vspace{-0.3in}
\label{fig:compare2}
\end{minipage}%
}
\subfigure[]{
\begin{minipage}[t]{0.3\linewidth}
\centering
\includegraphics[trim=-2cm -0.5cm 1.5cm 1cm, width=1.9in]{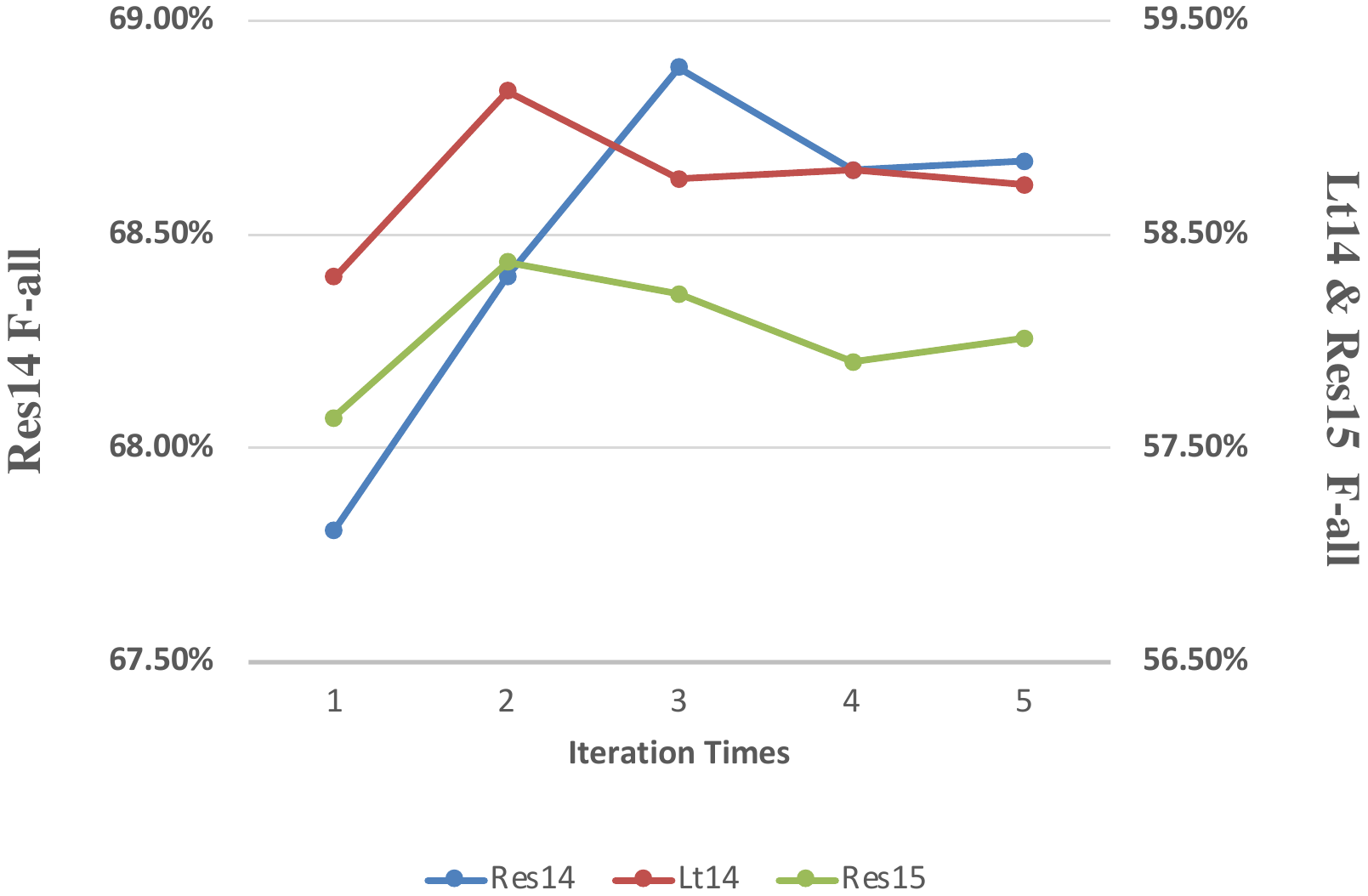}
\vspace{-0.3in}
\label{fig:compare3}
\end{minipage}%
}
\vspace{-0.1in}
\caption{Figure(a) shows comparison on multi-opinion-aspects instances between IMN and our Model. ``All'' denotes total test set and ``MultiAS'' denotes multi-opinion-aspects instances test set. Figure(b) shows comparison between \textbf{AE} model DECNN and our Model. ``NoAS'' means no-opinion-aspects instances test set. Figure(c) shows the F-all performance as iteration times increase.}
\label{fig:compare}
\end{figure*}

We compare our model with the state-of-the-art \textbf{TBSA} model \textbf{IMN}, which uses 30,000 external sentence-level sentiment data. The \textbf{F-all} performance in multi-opinion-aspects test set is shown in Figure~
\ref{fig:compare1}. We can observe that when the test set only contains the multiple-aspect instances, the performance gap between \textbf{IMN} and our model  is widened. Precisely, in Res14, Lt14 and Res15 multiple-aspect test sets, the gap is widened from 0.20\%, 1.57\% and 0.19\% to 2.49\%, 2.05\% and 0.46\%, respectively. This testifies that our model is indeed able to capture the relation among different aspects and sentiments towards them, thus improving the performance. 

\subsection{Performance on Instances Containing No Opinion Aspects}

In addition to the multi-opinion-aspects cases, we argue that without being aware of the sentiment information, the pipeline model would suffer from labeling many entities without sentiment polarities, which we call ``\textit{no-opinion-aspect}'' instances. In order to explore the effectiveness of our integration strategy on such instances, we select the instances which contain no opinion aspects from the test set. The statistics are shown in Table~\ref{tab:test_statistics}. We evaluate by \textbf{sentence-level accuracy} rather than F-a as the latter evaluation is always 0 in this case. The sentence-level accuracy metric is calculated as $\textit{acc} = \#\textit{(no-op\ instances)} / \#(\textit{all\ instances)} $.

We compare our model with the state-of-the-art \textbf{AE} model \textbf{DECNN}. The sentence-level accuracy performance in no-opinion-aspects test set is shown in Figure~\ref{fig:compare2}. The results show that our model beats the state-of-the-art {DECNN} by 2.06\%, 1.58\% and 3.17\% in three datasets, respectively. This indicates that without being aware of the sentiment information, the state-of-the-art {DECNN} model would suffer from labeling all the entities rather than opinion entities. Our model can ease this problem by considering the sentiment information in both implicit and explicit ways. More specifically, The implicit way we combine the sentiment information is the proposed shared structure that learns the shared hidden features. The explicit way of combining the sentiment is the word-sentiment edges in the proposed dynamic heterogeneous graph mechanism.

\subsection{Impact of Iteration Times of DHG}

We show the impact of iteration times on the {DHG} mechanism in Figure~\ref{fig:compare3}. We can observe that the \textbf{F-all} performance boosts as the iteration time increases by one or two, while begins to converge as iteration time further increases. This is a sign of the effectiveness of our DHG mechanism. From the results we also observe that the best number of iteration times is related to the ratio of multi-opinion-aspect instances (shown in Table \ref{tab:test_statistics}). To be more specific, the ratio of multi-opinion-aspect instances in Res14 dataset is higher than Lt14 and Res15 datasets, accordingly, we can observe that Res14 dataset needs more iterations than the others.

\begin{table*}
\centering
\scriptsize
\begin{tabular}{c|c|c|c}
\hline
Examples & IMN & Our Model\& Ground Truth & Prediction Process \\
\hline
\makecell[c]{I haven't used it for anything high tech \\yet, but I love it already.} 
& (tech)$_{\textit{pos}}$
& None 
& \makecell[c]{I haven't used it for anything high tech yet, but \\I love it already.} 
\\
\hline
\makecell[c]{I like coming back to Mac OS but \\this laptop is lacking in speaker quality \\compared to my \$400 old HP laptop.}
& 
\makecell[c]{(Mac OS)$_{\textit{neg}}$ \\ (speaker quality)$_{\textit{neg}}$}
&
\makecell[c]{(Mac OS)$_{\textit{pos}}$ \\ (speaker quality)$_{\textit{neg}}$}
&
\makecell[c]{
I$^{\color{red}\textit{pos}}$
like$^{{\color{red}\textit{pos}}}_{\color{red}\textit{pos}}$
coming$_{{\color{red}\textit{pos}}}^{\color{red}\textit{pos}}$
back$_{{\color{red}\textit{pos}}}^{\color{red}\textit{pos}}$
to$_{{\color{red}\textit{pos}}}^{\color{red}\textit{pos}}$
Mac$^{\color{red}\textit{pos}}$
OS$^{\color{red}\textit{pos}}$
but$^{\color{blue}\textit{neg}}$
this
\\[0.03in]
laptop
is$_{{\color{blue}\textit{neg}}}^{\color{blue}\textit{neg}}$
lacking$_{{\color{blue}\textit{neg}}}^{\color{blue}\textit{neg}}$
in$_{{\color{blue}\textit{neg}}}^{\color{blue}\textit{neg}}$
speaker$_{{\color{blue}\textit{neg}}}^{\color{blue}\textit{neg}}$
quality$^{\color{blue}\textit{neg}}$
\\[0.03in]
compared$^{\color{blue}\textit{neg}}$
to$^{\color{blue}\textit{neg}}$
my \$400 
old$^{\color{black}\textit{neu}}$
HP$_{{\color{black}\textit{neu}}}^{\color{black}\textit{neu}}$
laptop$_{{\color{black}\textit{neu}}}^{\color{black}\textit{neu}}$
.
\\[0.01in]
}
\\
\hline
\end{tabular}
\caption{Case Study between state-of-the-art IMN and our model. In the ``Prediction Process'', the subscript and the superscript denote the sentiment nodes which linked with word nodes of the first and second iteration, respectively.}\label{tab:casestudy}
\vspace{-0.12in}
\end{table*}

\subsection{Case Study}
We provide some concrete examples in the test set for case study in Table~\ref{tab:casestudy}. The first sample contains no aspect, as the opinion target is ``it'' rather than ``tech''. However, IMN predicts the word ``tech'' to be an aspect with positive polarity, which shows their superficial joint feature is not able to deal with such situations. In our model, DHG mechanism does not bridge the edge between ``tech'' and any sentiment labels. Our model successfully exclude the aspect ``tech'' which involves no opinion. 

The second sample contains two aspects with different sentiments. IMN model predicts both aspects as the same negative polarity, while our model distinguishes the different sentiments corresponding to different aspects. In the prediction process shown at the last column in the table, we can observe that DHG mechanism first builds the edges from sentiment nodes to opinion words (``like'', ``lacking'') and partial aspects (``speaker''), as they are predicted with high confidence. In the next iteration, the latter predictions of both aspect (``Mac OS'', ``quality'') and sentiment are aware of the previous opinion words and partial aspects. Moreover, our model observes that the word ``but'' reveals the opposite sentiment to the previous part ``I like coming back to Mac OS''.

\section{Related Work}

% aspect-level sentiment analysis
\subsection{Target-based Sentiment Analysis}

Target-Based Sentiment Analysis (\textbf{TBSA}) is an essential task in sentiment analysis and can be separated into two sub-tasks, which are aspect extraction (\textbf{AE}) and aspect sentiment analysis (\textbf{AS}). \textbf{AE} has been studied extensively by traditional machine learning~\citep{jakob2010extracting,liu2016improving} and deep neural network methods~\citep{xu2018double}. However, the absence of sentiment information results in redundant and noisy detection. \textbf{AS} aims to classify the sentiment expressed on some specific aspects in a sentence, which has been widely studied in the NLP community~\citep{wang-etal-2016-recursive,chen2017recurrent,ma2018targeted,chen2019transfer}. However, these aspects must be annotated before the \textbf{AS} task. \citet{wang2018towards} and \citet{li2019unified} link the two sub-tasks through unified tags. \citet{he2019interactive} fuse features of individual sub-tasks to common features via message-passing mechanism. However, these works remain on modeling the superficial connection between the two sub-tasks.

% graph Neural Network in Information Extraction
\subsection{GNN in Information Extraction}

Research on information extraction with Graph Neural Networks have been attracting heated attention. Advantages such as great power of expression, flexible design of structure, efficient encoding of knowledge base make GNN  achieved promising results in relation extraction~\citep{zhang2018graph}, event extraction~\citep{nguyen2018graph}, text classification~\citep{zhang2018sentence} and so on. Graph Convolutional Network~\citep{bruna2013spectral,duvenaud2015convolutional,kipf2016semi} , which generalizes the convolution operation from grid data to graph data, is one of the most popular variants of GNN. However, GCN is usually exposed to the over-smoothing problem when the number of layers increases~\citep{li2018deeper,chen2019measuring}. To relieve the over-smoothing, we use gate mechanism to filter the information to be transmitted to upper layers. 

\section{Conclusion}
This paper presents a novel integrated method for the \textbf{TBSA} task with dynamic heterogeneous graph, which explicitly models
the connection between text and sentiment. The graph is initialized with the syntactic and positional dependencies among words to model long distance dependencies. The combination of heterogeneous graph and various types of dependencies further builds path between keywords and sentiment labels.
Experiment results show that our model outperforms the state-of-the-art models. Detailed analysis reveals that both the syntactic dependency graph and the dynamic heterogeneous graph improve the performance. Moreover, our model is especially effective for detecting the connection among sentiments of different aspects  and excluding entities without opinions.

\bibliography{arxivjoint}
\bibliographystyle{acl_natbib}
\end{document}